\documentclass[conference]{IEEEtran}
\IEEEoverridecommandlockouts
\usepackage{cite}
\usepackage{amsmath,amssymb,amsfonts}
\usepackage{algorithmic}
\usepackage{graphicx}
\usepackage{textcomp}
\usepackage{xcolor}
\usepackage{siunitx}
\usepackage{url}
\def\BibTeX{{\rm B\kern-.05em{\sc i\kern-.025em b}\kern-.08em
    T\kern-.1667em\lower.7ex\hbox{E}\kern-.125emX}}
\begin{document}

% \title{Are Large Language Models Prepared to Serve as an End-to-End Compiler?}
\title{Exploring the Feasibility of End-to-End Large Language Model as a Compiler
\thanks{This work has been accepted by IJCNN 2025 and submitted to the IEEE for publication. Copyright may be transferred without notice, after which this version may no longer be accessible.}
}
% \thanks{This work was supported by the Strategic Priority Research Program of the Chinese Academy of Sciences, Grant No. XDA0320000 and XDA0320200}
% }

\author{
    Hongbin Zhang\textsuperscript{1,2},
    Shihao Gao\textsuperscript{1,2},
    Yang Liu\textsuperscript{1,2},
    Mingjie Xing\textsuperscript{1}\textsuperscript{*}
    \thanks{* Corresponding author: Mingjie Xing (mingjie@iscas.ac.cn)},
    Yanjun Wu\textsuperscript{1},
    Chen Zhao\textsuperscript{1}
    \\
    \textsuperscript{1}Institute of Software, Chinese Academy of Sciences, Beijing, China
    \\
    \textsuperscript{2}University of Chinese Academy of Sciences, Beijing, China
}

% \author{\IEEEauthorblockN{Anonymous Authors}}

% \author{\IEEEauthorblockN{1\textsuperscript{st} Given Name Surname}
% \IEEEauthorblockA{\textit{dept. name of organization (of Aff.)} \\
% \textit{name of organization (of Aff.)}\\
% City, Country \\
% email address or ORCID}
% \and
% \IEEEauthorblockN{2\textsuperscript{nd} Given Name Surname}
% \IEEEauthorblockA{\textit{dept. name of organization (of Aff.)} \\
% \textit{name of organization (of Aff.)}\\
% City, Country \\
% email address or ORCID}
% \and
% \IEEEauthorblockN{3\textsuperscript{rd} Given Name Surname}
% \IEEEauthorblockA{\textit{dept. name of organization (of Aff.)} \\
% \textit{name of organization (of Aff.)}\\
% City, Country \\
% email address or ORCID}
% \and
% \IEEEauthorblockN{4\textsuperscript{th} Given Name Surname}
% \IEEEauthorblockA{\textit{dept. name of organization (of Aff.)} \\
% \textit{name of organization (of Aff.)}\\
% City, Country \\
% email address or ORCID}
% \and
% \IEEEauthorblockN{5\textsuperscript{th} Given Name Surname}
% \IEEEauthorblockA{\textit{dept. name of organization (of Aff.)} \\
% \textit{name of organization (of Aff.)}\\
% City, Country \\
% email address or ORCID}
% \and
% \IEEEauthorblockN{6\textsuperscript{th} Given Name Surname}
% \IEEEauthorblockA{\textit{dept. name of organization (of Aff.)} \\
% \textit{name of organization (of Aff.)}\\
% City, Country \\
% email address or ORCID}
% }

\maketitle

\renewcommand{\thefootnote}{\fnsymbol{footnote}}

\begin{abstract}

% 近年来，端到端的大语言模型（LLM）技术在多个领域展现出显著优势。作为关键系统软件的编译器，其核心功能是将源语言转换为目标语言。虽然 LLM 技术已被应用于辅助编译器开发和维护，但将其直接用作端到端编译器的可能性尚未得到充分研究。本文率先探索了 LLM as a Compiler (LaaC) 的潜力与未来方向。我们设计了 CompilerEval 评估框架和数据集，专门用于测试主流 LLM 在源语言理解和汇编代码生成方面的能力。在评估过程中，我们以编译通过率为主要指标，深入分析了各类错误类型，并探索了多种提升编译通过率的方法。实验结果表明，LLM 确实具备理解源语言并生成跨平台汇编代码的能力，但编译通过率仍有较大提升空间。通过优化提示词、扩大模型规模和引入推理方法，编译通过率得到了显著提升。基于这些发现，我们对 LaaC 的前景持谨慎乐观态度，并提出了可行的架构设计和未来研究方向。我们相信，通过定向训练、专用知识库构建和基础设施的完善，LaaC 有望生成高质量汇编代码，并为编译领域带来范式转变。

In recent years, end-to-end Large Language Model (LLM) technology has shown substantial advantages across various domains. As critical system software and infrastructure, compilers are responsible for transforming source code into target code. While LLMs have been leveraged to assist in compiler development and maintenance, their potential as an end-to-end compiler remains largely unexplored. 
This paper explores the feasibility of LLM as a Compiler (LaaC) and its future directions. We designed the CompilerEval\footnote[2]{Open-source repository: \url{https://github.com/buddy-compiler/compiler-eval}} dataset and framework specifically to evaluate the capabilities of mainstream LLMs in source code comprehension and assembly code generation. In the evaluation, we analyzed various errors, explored multiple methods to improve LLM-generated code, and evaluated cross-platform compilation capabilities. Experimental results demonstrate that LLMs exhibit basic capabilities as compilers but currently achieve low compilation success rates. By optimizing prompts, scaling up the model, and incorporating reasoning methods, the quality of assembly code generated by LLMs can be significantly enhanced. Based on these findings, we maintain an optimistic outlook for LaaC and propose practical architectural designs and future research directions. We believe that with targeted training, knowledge-rich prompts, and specialized infrastructure, LaaC has the potential to generate high-quality assembly code and drive a paradigm shift in the field of compilation.
\end{abstract}


\section{Introduction}

% 编译技术自20世纪50年代中期以来经过广泛研究，已发展成为一个成熟的理论体系，其本质是将源语言转换为目标语言的过程。为确保转换的正确性和生成代码的质量，编译技术领域提出了一系列理论与方法。现代编译器大多采用三段式架构：前端、中端和后端。前端负责词法和语法分析，将源程序转换为token流，并构建抽象语法树或基于SSA的中间表示等数据结构。中端在此基础上执行语义分析和静态检查，同时进行机器无关的优化，如公共子表达式消除、常量折叠和循环不变量外提等。后端则专注于硬件相关的优化，最终生成目标代码。

\begin{figure}
    \centering
    \includegraphics[width=1\linewidth]{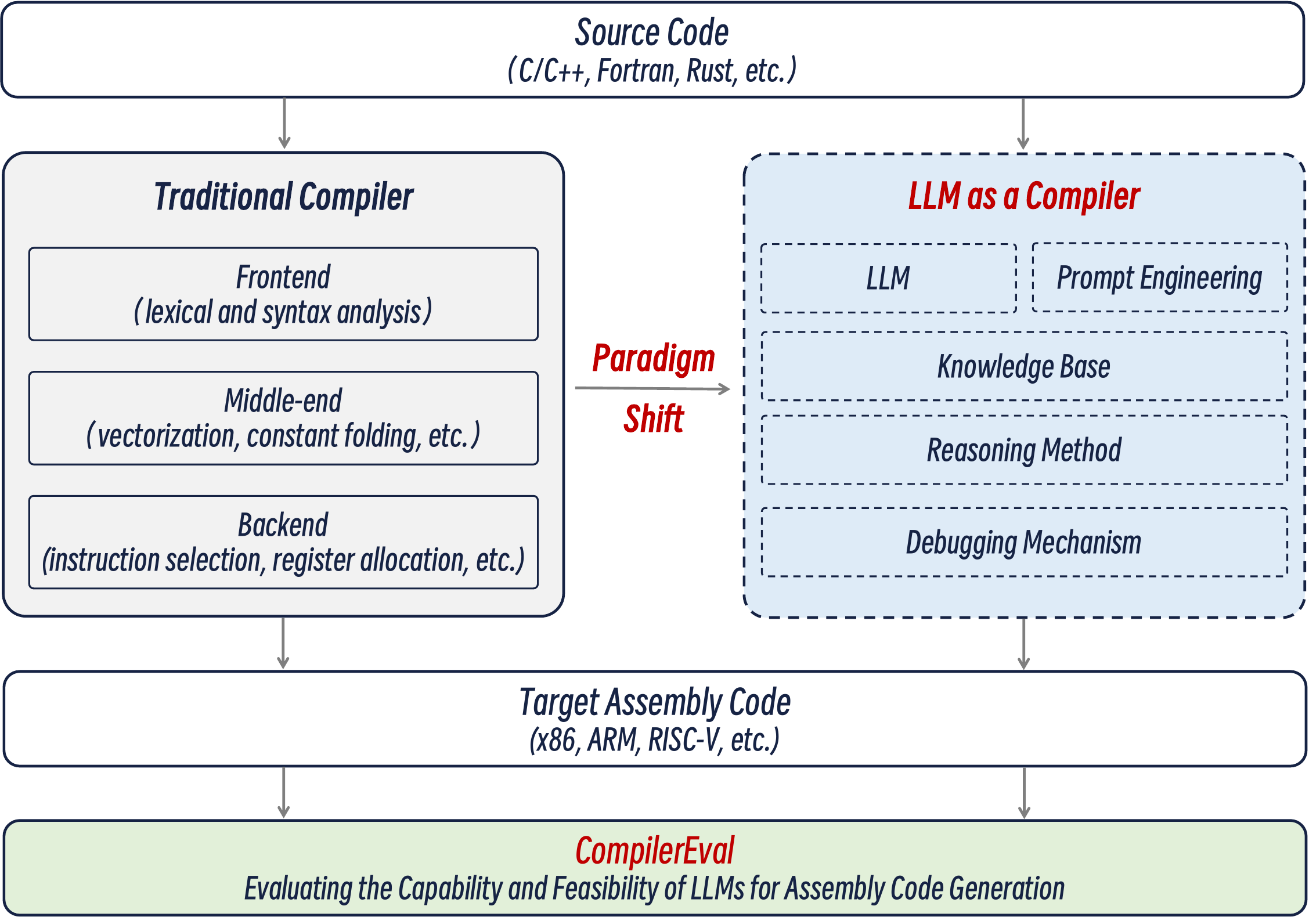}
    \caption{Paper Overview: Paradigm Shift from Traditional Compiler to LLM as a Compiler and Validation through the CompilerEval Dataset and Framework.}
    \label{fig:overview}
\end{figure}

Compiler technology has been extensively studied since the mid-1950s, developing into a well-established theoretical framework \cite{muchnick1997advanced,andrew2004modern,alfred2007compilers}. The purpose of a compiler is to transform source code into target code. As shown in Figure \ref{fig:overview}, modern compilers \cite{lattner2004llvm,gcc} generally adopt a three-phase architecture: frontend, middle-end, and backend. 
The frontend handles lexical and syntax analysis, converting source code into token streams and constructing intermediate representations (IRs). 
The middle-end performs semantic analysis and static checks while applying machine-independent optimizations. 
The backend focuses on hardware-specific optimizations and generates the target code.

% 编译器的复杂性导致了其高昂的开发和维护成本，如今编译器的全生命周期中越来越多地引入 AI 技术的辅助。在如今 DSL 和 DSA 蓬勃发展的时代，编译技术被广泛使用来解决语言和架构层面的碎片化问题，但是也增加了编译器的开发和维护成本。近年来，利用 AI 技术辅助编译器开发和维护已成为该领域的重要研究方向，主要聚焦于编译器优化、开发和测试三个核心领域。

% The complexity of compilers results in high development and maintenance costs, driving the increasing integration of artificial intelligence (AI) technologies throughout the compiler lifecycle. In the era of domain-specific languages and domain-specific architectures \cite{hennessy2019new,lattner2023golden}, compilation techniques \cite{MLIR,chen2018tvm,zhang2023compiler} are essential for addressing fragmentation at both the language and architectural levels. However, this also leads to significant human resource demands and manual costs. Consequently, leveraging AI methods to assist in compiler development and maintenance has emerged as a key research direction, with a focus on three core areas: optimization \cite{zhai2023tlp,mendis2019ithemal,turner2021neural,park2022srtuner,cummins2023large}, development \cite{zhongcomback,grossman2023compile}, and testing \cite{chakraborty2023ranking,munley2024llm4vv,yang2024whitefox}.

The complexity of compilers results in high development and maintenance costs, driving the increasing integration of artificial intelligence (AI) technologies throughout the compiler lifecycle. Compilation techniques \cite{MLIR,chen2018tvm,zhang2023compiler} are essential for addressing fragmentation at the language and architectural levels in the era of domain-specific languages and architectures \cite{hennessy2019new,lattner2023golden}. However, this also leads to significant human resource demands and manual costs. Consequently, 
compiler development and maintenance now utilize AI techniques for assistance, including optimization \cite{zhai2023tlp,mendis2019ithemal,turner2021neural,park2022srtuner,cummins2023large}, development \cite{zhongcomback,grossman2023compile}, and testing\cite{chakraborty2023ranking,munley2024llm4vv,yang2024whitefox}.

% 除了上述的 AI 辅助技术，本文致力于探索将 LLM 技术直接应用为 end-to-end 编译器的可能性。具体而言，我们设想将源代码作为输入直接提供给LLM，由其直接生成目标硬件平台的汇编代码。这一创新性构想源于 end-to-end LLM 技术在诸多领域的成功应用，包括自动驾驶、软件工程、语音识别和图像识别等。End-to-end LLM 的优势在于信息传递高效、系统设计简洁以及对复杂场景的强大适应能力。这些特性恰好能够解决当前编译器面临的关键挑战，如多级抽象导致的信息损失、模块设计的复杂性以及多硬件后端开发的冗余问题。若能成功将编译技术通过 end-to-end LLM 重新构建，将为计算机科学领域带来重要的范式转变。因此，我们认为 "LLM as a compiler" 代表了一个极具前景的研究方向。

This paper explores the feasibility of using end-to-end LLM technology as a compiler. Specifically, \textit{"end-to-end"} refers to the process where source code is input directly into the LLM, with the output being the assembly code for the target hardware platform. This idea is inspired by end-to-end LLM applications across various fields, such as autonomous driving \cite{yang2023llm4drive}, software engineering \cite{jin2023inferfix}, speech recognition\cite{ling2024adapting,lakomkin2024end}, and image recognition \cite{wang2024visionllm}. The advantages of end-to-end LLMs lie in their efficient information transfer, unified system design, and robust adaptability to complex scenarios. These characteristics are particularly well-suited to address the key challenges encountered by current compilers, including information loss in the lowering process, complexity in modular design, and redundancy in multi-hardware backend development. Therefore, we believe that LaaC represents a highly promising research direction. As shown in Figure \ref{fig:overview}, successfully reconstructing compiler technology with end-to-end LLMs could bring about a significant paradigm shift.

% 本文提出了 CompilerEval 框架和数据集，旨在全面评估当前 LLM 在汇编代码生成方面的能力。我们构建的数据集涵盖了线性代数、智能计算、图像处理和音频处理等多个关键领域，并采用四种主流 LLM 进行了系统性的测试验证。评估工作从三个维度展开深入分析：错误类型统计与修复策略、生成代码质量评估，以及模型规模与性能之间的关系。

This paper introduces the CompilerEval dataset and framework, which is designed to evaluate the capabilities of current LLMs in assembly code generation. The dataset covers 20 test cases across various domains, including linear algebra, intelligent computing, image processing, and audio processing. The framework integrates four mainstream LLMs as primary evaluation subjects, including OpenAI GPT \cite{GPT-4o,GPT-o1}, Google Gemini \cite{Gemini-2.0}, Anthropic Claude \cite{Claude-3.5}, and Meta Llama \cite{dubey2024llama}, while using \texttt{gcc} \cite{gcc} and \texttt{clang} \cite{lattner2004llvm} as benchmarks for correctness verification. The evaluation assesses the feasibility of LaaC from three perspectives: the basic ability of LLMs to generate assembly code, methods to enhance their capabilities, and the variations in cross-platform effectiveness.

% 研究结果表明，当前的 LLM 已具备生成可执行汇编代码的基本能力，通过精心设计的 prompt 能够产生正确的输出，这验证了 LLM 作为端到端编译器的潜力。然而，要实现稳定可用的 LLM 编译器仍面临诸多挑战。通过深入的数据分析，我们将这些挑战归纳为三个核心问题：（1）代码正确性保证不足；（2）模型训练难度较大；（3）决策过程可解释性欠佳。针对这些挑战，本文提出了一系列未来研究方向，包括高质量数据集构建、先进学习技术应用、创新网络架构设计以及领域知识深度融合等。

The experiment results show that mainstream LLMs possess the basic capability to generate executable assembly code, but the average compilation success rate remains low. The quality of the LLM-generated code can be improved by applying prompt engineering, scaling up the model, and incorporating reasoning methods. Additionally, LLMs with these mechanisms can support cross-platform compilation requirements. Through in-depth data analysis, this paper proposes the LaaC design and outlines three key research directions: (1) training LLMs to meet compilation scenario constraints, (2) developing the infrastructure of LaaC to support multi-language to multi-platform compilation, and (3) aligning LaaC and debugger designs to enhance debugging efficiency.

% 综上所述，本研究的主要创新在于首次系统性地探索了将 end-to-end LLM 技术应用为编译器的可能性，并通过实验论证了其可行性。

% 本研究的主要贡献如下：

% - 设计并实现了 CompilerEval 评估框架，同时构建了专门的测试数据集，为验证 LLM 在编译领域的应用能力提供了重要工具。
% - 开展了全面的评估与深入分析，涵盖错误类型统计与修复策略、生成代码质量评估，以及模型规模与性能关系等多个维度。
% - 明确指出了未来关键研究方向，包括模型精调策略与prompt工程优化、debug 能力增强，以及模型规模与性能的平衡等重要研究点。

In summary, the primary innovation of this research is the exploration of applying end-to-end LLM technology as a compiler and demonstrating its feasibility through experimental validation.

The main contributions of this research are as follows:

\begin{enumerate}
    \item We designed and implemented the CompilerEval dataset and framework to evaluate the capabilities of LLMs in the compiler domain.
    \item We performed evaluation and analysis on various mainstream LLMs, specifically focusing on assessing compilation capabilities and limitations, improving compilation success rates, and generating cross-platform assembly code.
    \item We identified key design concepts and future research directions for LaaC, including LLM training, prompt engineering, chain-of-thought reasoning, specification knowledge base, and intelligent debugging mechanisms.
\end{enumerate}

% 本文的结构安排如下：第二章详细阐述 CompilerEval 的设计理念、框架结构及数据集构建；第三章系统性地展示评估结果并进行深入分析；第四章探讨 LLM 作为端到端编译器面临的挑战及未来研究方向；第五章综述相关研究工作；第六章对全文进行总结。

The paper is structured as follows: Section II presents background and motivation; Section III details the CompilerEval design; Section IV provides a systematic analysis of the evaluation results; Section V discusses the challenges and future research directions of LaaC; Section VI reviews related work; Section VII concludes the paper.

\begin{figure*}
    \centering
    \includegraphics[width=0.8\linewidth]{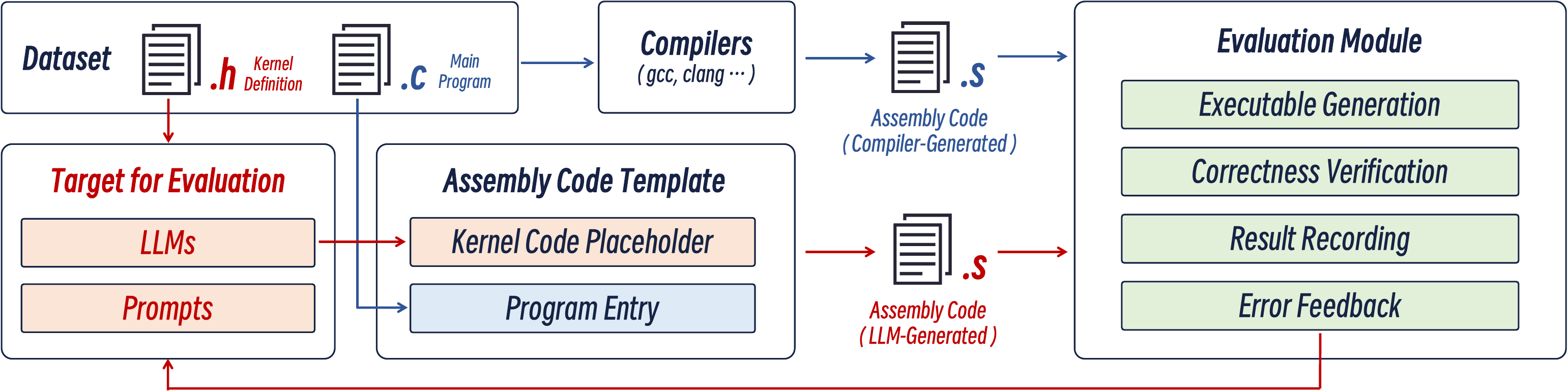}
    \caption{Overview of the CompilerEval Framework.}
    \label{fig:compilar_eval}
\end{figure*}

\section{Background and Motivation}

% 在生成式人工智能技术日益应用于软件基础设施的背景下，本研究探索大语言模型（LLM）直接将源代码转换为汇编代码的潜力，并评估这一技术能否在编译领域引发设计范式的转变，以此解决传统编译器设计与实现中的固有问题。

% Motivated by the growing adoption of generative AI in software infrastructure, this research explores the feasibility of LLMs to directly compile source code into assembly code. This section discusses whether this novel approach could fundamentally transform paradigms and address challenges in traditional compiler design and implementation.

This section discusses the background of compilation principles and highlights the motivation for using LLMs to compile source code directly into assembly code. With the growing adoption of generative AI in software infrastructure, we expect to determine whether this novel approach could address challenges in traditional compiler design and implementation.

% 传统编译器经过70年左右的发展，整体架构具有很强的理论性和鲁棒性，但如今仍然面临一系列挑战，本文将其总结为以下三个大类：（1）前后端通用性不足 - 尽管已有众多模块化和接口优化的研究，但支持新的编程语言和硬件架构时仍需大量人力和精力进行适配；（2）中端信息传递损耗 - 即便采用了多级中间表示和符号化形状等技术，编译过程中的抽象变换仍不可避免地导致信息损失，从而影响编译优化效果；（3）系统复杂度高 - 尽管已经有一些研究在使用 AI 辅助编译器开发和维护，但编译器仍是一个需要深厚专业知识的精密系统，具有较高的开发难度和学习门槛。

Although compilation principles have developed over 70 years, the design and implementation of compilers still face significant challenges, which we identify as three main areas: 
\textbf{(1) Limited frontend and backend generality}. 
Although modular designs have been studied \cite{parr1995antlr,lattner2004llvm}, supporting new programming languages and hardware architectures requires significant human effort.
\textbf{(2) Information loss in middle-end processing}. Even with multi-level IRs \cite{MLIR}, information loss during the compilation process remains inevitable, affecting the effectiveness of optimizations.
\textbf{(3) High system complexity}. Although AI-assisted approaches have been explored \cite{zhai2023tlp,mendis2019ithemal,turner2021neural,park2022srtuner,cummins2023large,zhongcomback,grossman2023compile,chakraborty2023ranking,munley2024llm4vv,yang2024whitefox}, compilers remain complex systems requiring deep expertise, 
with high development difficulty and steep learning curves.

% 端到端 LLM 的技术特点为解决上述编译领域的问题提供了新的可能性，这也是本文提出将 LLM 作为编译器进行深入研究的核心动机。LLM 具有以下三个显著优势：（1）强大的新任务适应能力：在面对新的编程语言和硬件指令时，端到端 LLM 展现出强大的通用性和适应性。通过特定的知识库和 prompts，它能够快速适应不同场景，无需针对每个语言或硬件进行大量的定制化开发，显著降低了开发成本。（2）高效的信息传递机制：端到端 LLM 通过直接映射的方式，避免了传统编译过程中多次转换导致的信息损失和误差累积。在海量数据的训练支持下，模型能够深入理解源语言和目标语言之间的本质联系，从而有潜力生成更优质的汇编代码。（3）简洁的系统架构：端到端 LLM 突破了传统编译器复杂的多模块设计范式，将整个编译流程融入统一的模型框架中。这种设计极大地简化了系统设计和实现的复杂性，减少了模块间协调和集成的工作量。

The characteristics of end-to-end LLM techniques present new opportunities for tackling the challenges mentioned above, which forms the core motivation of this paper. We identify three key advantages that make them well-suited for the compilation domain: 
\textbf{(1) Strong adaptability to new tasks}. LLMs exhibit strong generality and adaptability by leveraging specialized knowledge bases and prompts. This capability offers an efficient and unified approach for supporting new programming languages and hardware architectures, reducing the development cost in compiler frontend and backend implementation. 
\textbf{(2) Efficient information transmission mechanism} - By direct mapping, LLMs avoid information loss and error accumulation caused by multiple transformations in traditional compilation processes. With support from massive training data, these LLMs can grasp the fundamental relationships between source and target languages, offering the potential to generate higher-quality assembly code. 
\textbf{(3) Unified system architecture} - LLMs can integrate the entire compilation process into a unified model framework, significantly simplifying compiler system design and implementation. This approach reduces the complexity of coordination and integration across individual modules.

\section{CompilerEval Dataset and Framework}

% 本文提出了 CompilerEval 用于验证 LLMs 接受源代码并直接生成汇编代码的能力，从而给 LaaC 的设计和可行性分析提供参考。本章节详细阐述 CompilerEval 的数据集构建和框架架构。

% This paper introduces CompilerEval to assess LLMs' ability to take source code as input and directly generate assembly code, providing valuable insights for the design and feasibility analysis of LaaC. This section details the dataset construction and framework architecture of CompilerEval.

This section introduces CompilerEval, a framework that evaluates the ability of LLMs to take source code as input and generate assembly code directly. The following part details the dataset construction and the architectural design of CompilerEval.

\begin{table}[ht]
\caption{Test Cases in the CompilerEval Dataset}
\small
\begin{tabular}{p{1.25cm} p{6.75cm}}
\hline
\textbf{Name}         & \textbf{Description}                           \\ \hline
trmm         & Triangular matrix-matrix multiplication.   \\
gemver          & General matrix-vector multiplication. \\
gesummv         & Generalized matrix-vector multiplication.   \\
2mm        & Matrix-matrix multiplication.    \\
mvt         & Matrix-vector transpose multiplication. \\
saxpy    & Scalar-vector addition.     \\
sgemm       & Single-precision matrix-matrix multiplication.      \\
conv2d       & 2D convolution operation.     \\
softmax   & Activation function for classification.    \\
pooling   & Downsampling operation in CNNs.     \\
relu   & Rectified Linear Unit activation.   \\
resize   & Image resizing.     \\
rotate   & Image rotation.     \\
fir   & Finite Impulse Response filter.      \\
iir   & Infinite Impulse Response filter.  \\
correlation   & Measure of the relationship between two variables.        \\
covariance   & Measure of variance between variables.        \\
fdtd-2d   & 2D Finite Difference Time Domain simulation.   \\
jacobi-1d   & Jacobi method for solving 1D linear equations.       \\
jacobi-2d   & Jacobi method for solving 2D linear equations.        \\
\hline
\end{tabular}
\label{tab:dataset}
\end{table}

As illustrated in Table \ref{tab:dataset}, the CompilerEval dataset consists of 20 representative cases from various domains, including linear algebra, intelligent computing, image processing, and audio processing.
Each case in the dataset is written in C and divided into the kernel definition and the main program.

% CompilerEval 框架旨在通过对比编译器和 LLM 生成的汇编代码来评估其质量和正确性。如图所示，该框架由数据集、评估目标、汇编代码模板、编译器和评估模块等核心组件构成。评估过程主要分为两个部分：

The CompilerEval framework is designed to evaluate the quality and correctness of assembly code by comparing outputs generated by traditional compilers and LLMs. 
As illustrated in Figure \ref{fig:compilar_eval}, the components of the framework include a dataset, evaluation targets, assembly code templates, compilers, and evaluation modules.
The evaluation process is primarily divided into two parts:

\textbf{(1) Assembly code generation.} 
As shown on the left of Figure \ref{fig:compilar_eval}, LLMs convert kernel definitions into assembly code using prompts, which are then inserted into predefined templates.
Meanwhile, the main program from the dataset is integrated into the template as the program's entry point, completing the assembly code.
As a baseline, traditional compilers generate assembly from the same source code to verify the correctness and quality of the LLM-generated output.

\textbf{(2) Assembly code evaluation.}
The right side of Figure \ref{fig:compilar_eval} shows the evaluation module, which includes four steps: executable generation, correctness verification, result recording, and error feedback. 
Equation \ref{eq:metrics} defines $\texttt{success@1}$ as the metric for \textit{Compilation Success Rate}, with $N_{\text{total}}$ = 200 in this paper.
Additionally, we use \textit{Executable Samples Count} ($N_{\text{exec}}$) to represent lexical and syntactic correctness and \textit{Success Samples Count} ($N_{\text{succ}}$) to represent semantic correctness. 
The \textit{Executable Samples Correctness Rate} ($R_{\text{exec\_corr}}$) is used to measure the quality of executable samples. 
% Finally, CompilerEval collects results and errors into an evaluation report.

\begin{equation}
\text{success@1} := 1 - \frac{\binom{N_{\text{total}}-N_{\text{succ}}}{1}}{\binom{N_{\text{total}}}{1}}, \quad R_{\text{exec\_corr}} := \frac{N_{\text{succ}}}{N_{\text{exec}}}
\label{eq:metrics}
\end{equation}

\section{Evaluation Results}

% 本章节深入探究 LaaC 的潜力。我们将从多个维度进行评估：首先统计各个 LLM 在 CompilerEval 上的编译通过率与各类编译错误，然后评估多种主流提高 LLM 生成效果的方法，最后评估 LLM 跨硬件平台生成汇编代码的能力。

This section evaluates the feasibility of LaaC using CompilerEval from three perspectives: (1) analyzing the assembly code generation capabilities and limitations, (2) evaluating methods to improve compilation success rates, and (3) assessing the ability to generate cross-platform assembly code.

\subsection{Evaluation of Assembly Code Generation Capabilities and Limitations}

\begin{figure}
    \centering
    \includegraphics[width=1\linewidth]{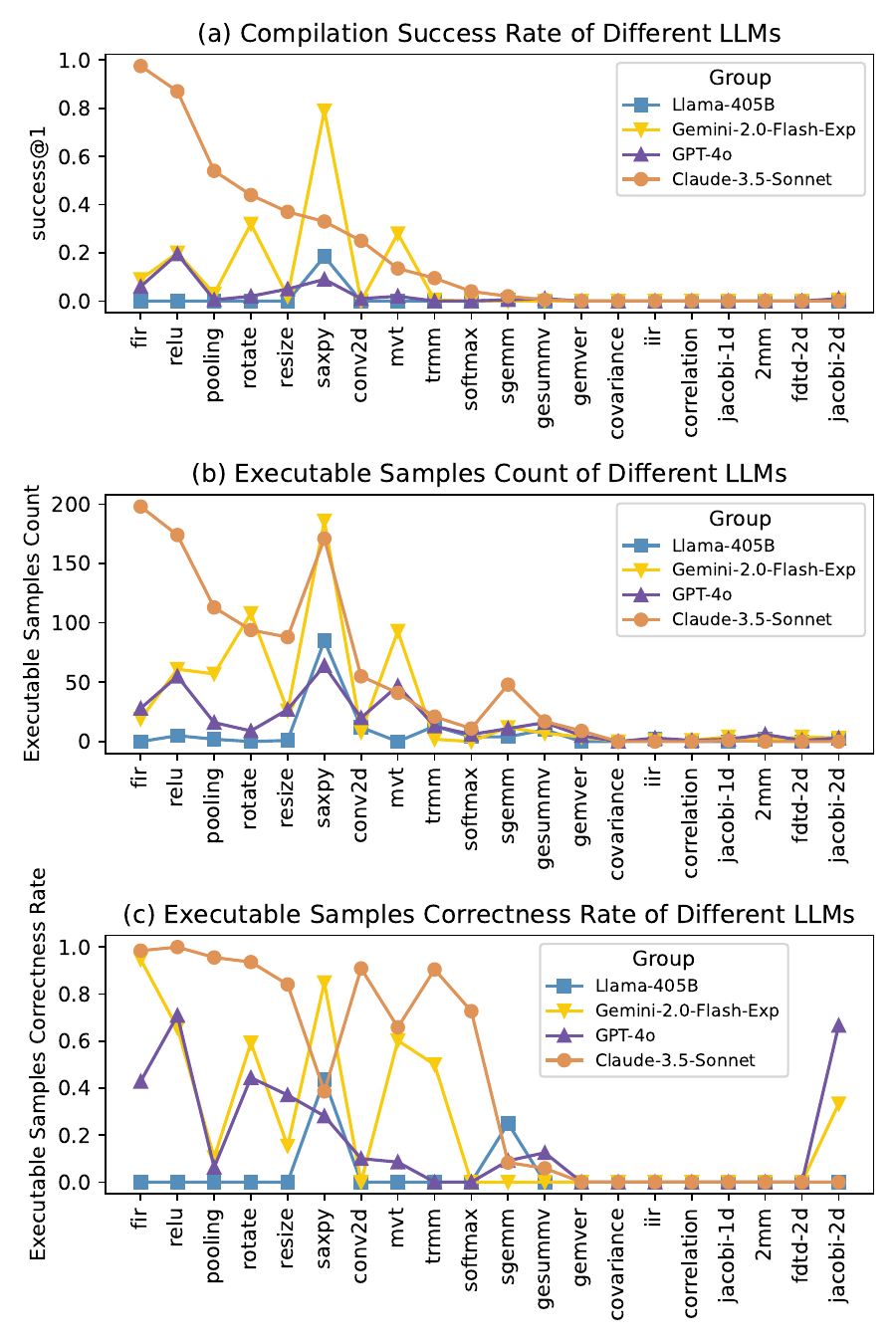}
    \caption{Results and Trends of Mainstream LLMs on the CompilerEval Dataset.}
    \label{fig:llms}
\end{figure}

% 本节实验评估了 LLM 在汇编代码生成方面的基础能力。我们重点分析了编译通过率这一核心指标，并对编译错误进行了系统性的统计与分类。实验选取了四个代表性的大语言模型：GPT-4o、Claude-3.5-Sonnet、Gemini-2.0-Flash-Exp 和 LLaMA-3.1-405B。我们首先在 x86 平台上开展实验，通过收集和分析编译成功率及错误数据，进一步讨论了当前主流 LLM 在汇编代码生成任务中的表现与局限。

This experiment evaluated the basic capabilities of LLMs in assembly code generation by analyzing compilation success rates and errors. We selected four mainstream LLMs: GPT-4o \cite{GPT-4o}, Claude-3.5-Sonnet \cite{Claude-3.5}, Gemini-2.0-Flash-Exp\cite{Gemini-2.0}, and LLaMA-3.1-405B\cite{dubey2024llama}. This experiment was conducted on the x86 platform, and the results provided insights into the performance and limitations of current mainstream LLMs in assembly code generation.

% 在汇编代码生成能力方面，不同 LLM 表现差异显著。图 XX (a) 和 (b) 四种模型在不同用例上的编译效果差异明显，仅在 saxpy 用例的可执行样本数量上呈现一致趋势。这反映出不同模型在处理语法结构时各有优势，Claude-3.5-Sonnet 表现最为突出, 而所有模型在处理复杂用例时均表现不佳。图 XX-(c) 对比了可执行样本的正确率，Claude-3.5-Sonnet 仍然表现最优。这表明其生成的汇编代码通过词法和语法检查后，语义正确性也较高。相比之下，其他模型生成的可执行程序往往难以保证语义正确性。

Different LLMs show significant variations in their assembly code generation capabilities. 
Figures \ref{fig:llms} (a) and (b) reveal variations in compilation effectiveness among the four models across different test cases. Notably, only the \texttt{saxpy} test case shows consistency in the number of executable samples.
This indicates that each model has its own strengths in handling syntactic structures, with Claude-3.5-Sonnet showing the most outstanding performance. 
However, all models struggle with complex test cases. 
Figure \ref{fig:llms}-(c) compares the executable samples correctness rates, where Claude-3.5-Sonnet still performs the best.  
This indicates that when Claude’s generated assembly code passes lexical and syntactic validation, it tends to maintain high semantic correctness as well.
In contrast, the executable programs generated by other models often fail to ensure semantic correctness.

% LLM 生成的汇编代码存在多种问题。以表现最佳的 Claude-3.5-Sonnet 为例，图 XX 展示了编译错误、执行错误和正确执行的分布。从图中红色部分可见，编译期错误占比最高，这表明了即使是效果最好的 Claude-3.5-Sonnet 也难以保证汇编代码的语法正确性。我们进一步分析了 4 种模型在 CompilerEval 上的错误信息，将其归为 9 类，统计结果见表格 XX。数据显示 LLMs 在生成汇编代码时存在大量词法语法问题，链接错误反映了 LLMs 符号处理能力不足，运行时错误则主要是内存访问问题，如段错误和结果错误等。

\begin{figure}
    \centering
    \includegraphics[width=1\linewidth]{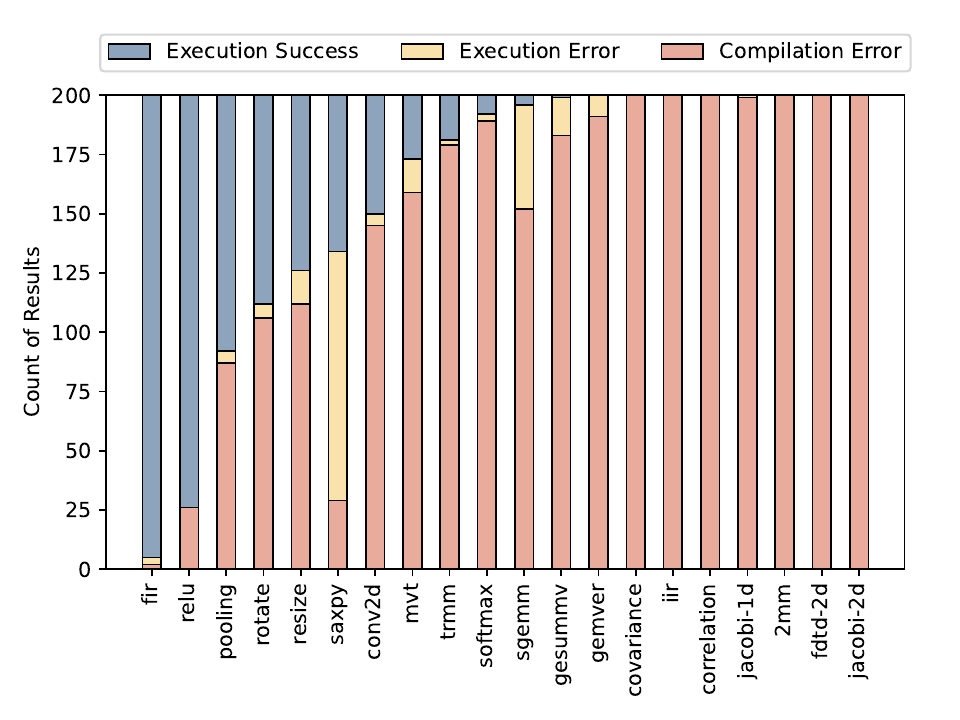}
    \caption{Distribution of Execution Success, Execution Errors, and Compilation Errors for Claude-3.5-Sonnet on the CompilerEval Dataset.}
    \label{fig:sample-status}
\end{figure}

The assembly code generated by LLMs exhibits several issues. Taking the best-performing model Claude-3.5-Sonnet as an example, Figure \ref{fig:sample-status} shows the distribution of successful executions, execution errors, and compilation errors. As shown by the red portion in Figure \ref{fig:sample-status}, compilation errors account for the largest proportion, indicating that even Claude-3.5-Sonnet struggles to ensure syntactic correctness. The yellow portion in Figure \ref{fig:sample-status} represents execution errors, with the \texttt{saxpy} test case showing the highest number of errors due to incorrect register and program variable binding. We further analyzed error messages from four models on CompilerEval, categorizing them into nine types. As shown in Table \ref{tab:error}, the data reveals that LLMs encounter numerous lexical and syntactic issues when generating assembly code, primarily including instruction errors, invalid register usage, and incorrect symbol handling. Execution errors mainly involve memory access issues, such as segmentation faults and wrong results.

% 总体而言，实验结果表明 LLM 具备从源语言生成汇编代码的基本能力。然而，当前仅能将较简单程序正确编译并执行，整体编译通过率较低。这意味着，在现阶段，直接依赖通用 LLM 生成汇编代码仍缺乏实用价值。

In summary, the experimental results indicate that LLMs possess the basic capability to generate assembly code from source code. However, they are limited to correctly compiling and executing simple programs, with a low overall compilation success rate. This suggests that relying solely on general-purpose LLMs for assembly code generation lacks practical value at the current stage.

\begin{table}[ht]
\caption{Error Analysis \\
Compilation and Execution Errors with Occurrence Counts}
\small
\begin{tabular}{p{4cm} p{2.5cm} p{1cm}}
\hline
\textbf{Error Name}             & \textbf{Error Category}     & \textbf{Count}    \\ \hline
Unrecognized Character    & Compilation Error & \num{2135}      \\
Instruction Error         & Compilation Error & \num{7269}      \\
Absolute Expression Error & Compilation Error & \num{1880}      \\
Invalid Register Usage    & Compilation Error & \num{2052}      \\
Undefined Reference Error & Compilation Error      & \num{267}      \\
Symbol Redefinition Error & Compilation Error      & \num{2556}      \\
Segmentation Fault        & Execution Error     & \num{1156}      \\
Illegal Instruction       & Execution Error     & \num{438}      \\
Wrong Result    & Execution Error     & \num{622}      \\
\hline
\end{tabular}
\label{tab:error}
\end{table}

\subsection{Evaluation of Methods to Improve Compilation Success Rates}

\begin{figure*}
    \centering
    \includegraphics[width=1.0\linewidth]{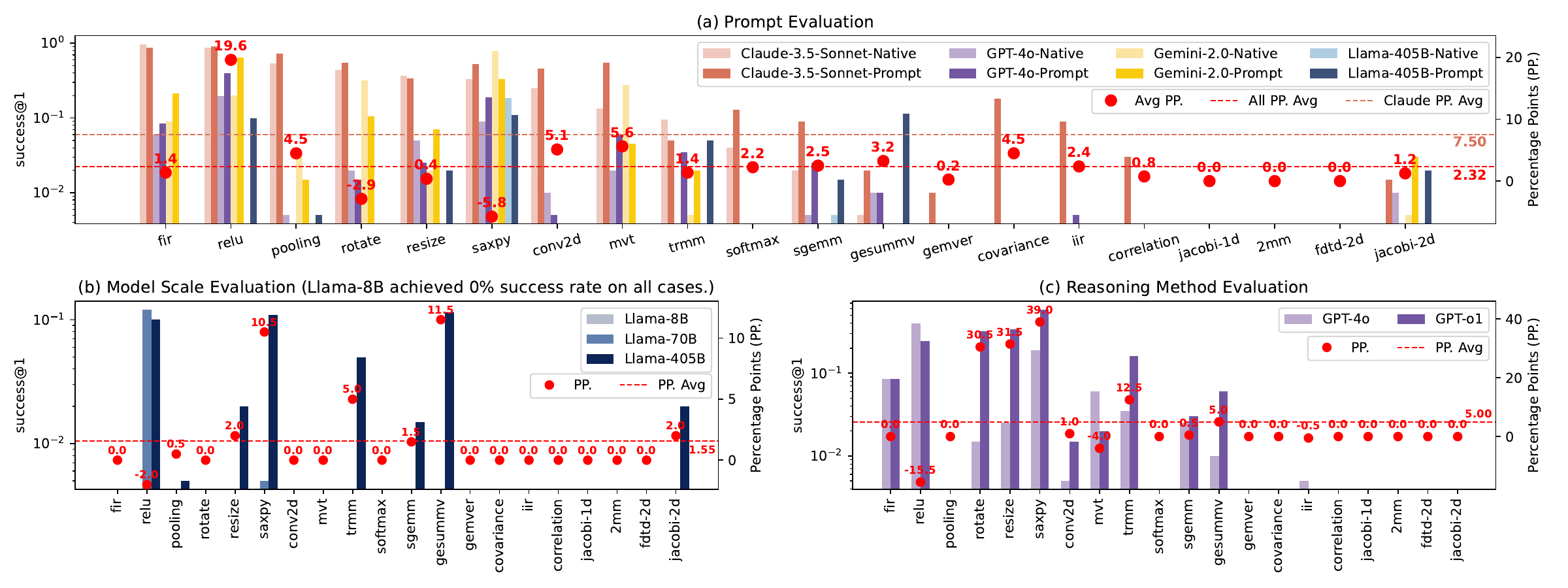}
    \caption{Evaluation of Prompting Engineering, Model Scale, and Reasoning Method on LLM-based Compilation Success Rate.}
    \label{fig:compare}
\end{figure*}

% 本实验评估了多种提升 LLM 汇编代码生成能力的方法，包括 prompt engineering、模型规模和 reasoning 方法对编译通过率的影响。实验主要通过分析编译通过率的提升百分点来评估各方法的有效性。如图 XX 所示，我们使用对数坐标系的柱状图展示了各方法的 success@1 数值，并用散点表示每组实验的平均提升百分点，虚线则代表所有散点的平均值。通过虚线处的数值，我们可以直观地判断不同方法对提升编译通过率的效果。

This experiment evaluated several approaches to enhance LLMs' assembly code generation, focusing on the impact of prompt engineering, model scale, and reasoning methods. The effectiveness is measured through the percentage point increase in compilation success rates. Figure \ref{fig:compare} presents the \texttt{success@1} values using a log-scale bar chart with scatter points showing group averages. A dashed line indicates the overall average, making it easy to illustrate the effectiveness of each method.

% 为评估 prompt 工程对编译通过率的影响，我们通过分析错误类型设计了针对性的 prompt，并对比了添加与不添加 prompt 的编译通过率。如图 XX-(a)所示，大多数情况下添加 prompt 能够提升通过率，其中 Claude 模型的提升最为显著，达到了 7.5 个百分点，而 Llama 的改善效果较小。然而，在部分案例中也观察到了负优化现象，特别是在 Gemini 模型处理 pooling、rotate、saxpy 和 mvt 等任务时。综合所有模型的表现，添加 prompt 后的平均编译通过率提升了2.32个百分点。

To assess the impact of prompt engineering, we designed targeted prompts for each type of error in Table \ref{tab:error}. As shown in Figure \ref{fig:compare}-(a), adding prompts generally improved the success rate, with Claude demonstrating the largest increase of 7.5 percentage points, while Llama showed minimal gains. However, some negative effects were observed, particularly in Gemini's performance on \texttt{pooling}, \texttt{rotate}, \texttt{saxpy}, and \texttt{mvt} kernels. Across all models, prompts improved the average compilation success rate by 2.32 percentage points.

% 实验使用不同规模的 Llama 模型 8B、70B和 405B 参数来评估模型大小对编译通过率的影响。如图 XX-(b) 所示，随着模型规模的增加，编译通过率呈现明显的提升趋势。具体而言，Llama-8B 在所有测试用例中均未能生成任何可成功编译和正确执行的代码。Llama-70B 展现出了初步的代码生成能力，成功在 relu 和 saxpy 两个样例上生成了正确的汇编代码。当扩展到 Llama-405B 时，可以正确生成的样例数量显著增加。图 XX-(b) 中的散点表示 Llama-405B 相对于 Llama-70B 提升的百分点，在所有样例中平均提升了 1.55 个百分点的通过率，其中在 saxpy 和 gesummv 样例下的提升甚至超过了 10 个百分点。

This experiment evaluated model scaling effects using Llama variants of 8B, 70B, and 405B parameters. 
Figure \ref{fig:compare}-(b) reveals a clear correlation between model size and compilation success rates. 
Specifically, Llama-8B failed to generate any executable code; 
Llama-70B successfully produced correct assembly for \texttt{relu} and \texttt{saxpy} tasks; 
Llama-405B significantly improved performance.
The scatter point in \ref{fig:compare}-(b) indicates that Llama-405B achieves an average improvement of 1.55 percentage points in success rate compared to Llama-70B.
The most notable gains are observed in \texttt{saxpy} and \texttt{gesummv} tasks, exceeding 10 percentage points.

% 实验评估了 reasoning 方法对编译通过率的影响，对比了 GPT-4o 和具有 reasoning 能力的 GPT-o1 两个模型。如图 XX-(c) 所示，在大多数情况下，具备 reasoning 能力的 GPT-o1 表现优于 GPT-4o。特别是在 rotate、resize 和 saxpy 等案例中，编译通过率提升显著，超过了 30 个百分点。虽然在 relu 和 mvt 等少数案例中 GPT-o1 的表现略逊，且对于较复杂的程序仍无法生成正确的汇编代码，但整体而言，使用具有 reasoning 能力的模型平均带来了 5 个百分点的通过率提升。

We examined the impact of reasoning methods on compilation success rates by comparing GPT-4o with GPT-o1 \cite{GPT-o1}, a model equipped with reasoning capability. Figure \ref{fig:compare}-(c) demonstrates that GPT-o1 achieved superior performance across most test cases. Significant improvements exceeding 30 percentage points were observed in \texttt{rotate}, \texttt{resize}, and \texttt{saxpy} kernels. Although GPT-o1 showed slightly lower performance in some cases (\texttt{relu} and \texttt{mvt}) and faced challenges with complex programs, the reasoning-enhanced model achieved an overall improvement of 5 percentage points in compilation success rates.

% 总体而言，我们的实验表明添加 prompt、扩大模型规模以及引入 reasoning 方法都能提升 LLM 在汇编代码生成方面的性能。尽管这些方法在本实验中并未针对编译场景进行深度优化，导致提升幅度有限，但实验结果仍为未来研究提供了重要启示。我们认为，通过将 LLM 与编译领域特定的优化方法相结合，未来有望实现高质量的汇编代码生成。

In summary, our experiments show that the effectiveness of LLMs in generating assembly code benefits from advances in prompt engineering, model scaling, and reasoning capabilities.
% In summary, our experiments show that prompt engineering, model scaling, and reasoning capabilities each contribute to enhancing LLMs' assembly code generation effectiveness. 
While these approaches yielded limited improvements in our exploration, they provide valuable insights for future research directions. 
The findings suggest that integrating LLMs with compilation-specific assistant techniques could potentially enable high-quality assembly code generation in future applications.

\subsection{Evaluation of Cross-Platform Assembly Code Generation Capabilities}

% 跨硬件平台代码生成是编译器的基本能力之一，本实验专门评估了 LaaC 在跨平台汇编代码生成方面的性能表现。根据前期实验数据，我们选择了编译通过率最优的 Claude 模型进行深入研究，并在 x86、ARM 和 RISC-V 这三个主流指令集架构上进行了全面测试。

The ability to generate cross-platform assembly code is essential for modern compilers. 
This experiment evaluated LLMs' capability in cross-platform assembly generation, focusing on the Claude-3.5-Sonnet model, which achieved the highest compilation success rates in previous tests. 
We conducted comprehensive assessments across three mainstream instruction set architectures: x86, ARM, and RISC-V.

\begin{table}[ht]
\caption{Hardware Information and Compilation Success Rates}
\begin{center} % 使表格居中
\begin{tabular}{c c c c}
\hline
Machine         & Arch.  & Device Type & success@1 Avg. \\ \hline
Xeon Gold 5218R & x86    & Server  &   27.85\%  \\
Apple M1 Max    & ARM    & Laptop   &  35.02\%  \\
SpacemiT K1     & RISC-V & Embedded Board  &   32.30\%  \\  \hline
\end{tabular}
\end{center}
\label{tab:hardware-info}
\end{table}

% 首先，实验将 CompilerEval 部署到多个硬件平台并统计编译通过率。如图表所示，我们的测试设备类型包括服务器、笔记本电脑、和嵌入式板卡。得益于 LLM 基础设施的完备，我们的实验可以通过 LLM API 在各种类型的设备上调用。实验中，我们使用 Claude 分别面向三种架构生成汇编代码。实验结果表明，Claude 这样的 LLM 有能力向不同架构生成正确的汇编代码。我们进一步统计了每种架构上的编译通过率的平均值。如图所示，实验结果表明，ARM 和 RISC-V 拥有比 x86 更好的编译通过率，其中 ARM 的平均通过率最高，RISC-V 比 ARM 低了2.72个百分点，比 x86 高了 4.45 个百分点。

This experiment deployed CompilerEval across multiple hardware platforms to measure compilation success rates. As shown in Table \ref{tab:hardware-info}, we tested on servers, laptops, and embedded boards, and the results confirmed that LLMs like Claude can generate correct assembly code across architectures. Analysis of compilation success rates revealed that both ARM and RISC-V architectures achieved higher rates than x86. The results of ARM demonstrated the highest average success rate, while RISC-V's rate was 2.72 percentage points below ARM's but 4.45 percentage points above x86's.

% 进一步，我们统计了实验中可执行案例的数量，并分析了其执行正确率，以此全面评估各架构汇编代码生成的质量。如图所示，散点图中的每个点表示在对应硬件架构上，200个测试样例中可以正确执行的数量。每一行的散点图数据代表 CompilerEval 数据集中的 20 个 workload。在每一行中，散点越靠右说明生成的汇编代码语法错误越少，可以生成可执行文件。图中还展示了每个架构的可执行样例数量平均值，实验结果表明 RISC-V 后端拥有最多的可执行样例，ARM 次之，而 x86 的可执行样例数量最少。此外，散点的颜色深浅表示该测试用例执行正确率的高低。因此，右侧的深色散点越多，说明编译效果越好。从图中可以观察到，ARM 和 RISC-V 架构在右侧的深色散点数量显著多于 x86，这表明在 Claude 生成的 ARM 和 RISC-V 的汇编代码质量明显优于 x86。

\begin{figure}
    \centering
    \includegraphics[width=1\linewidth]{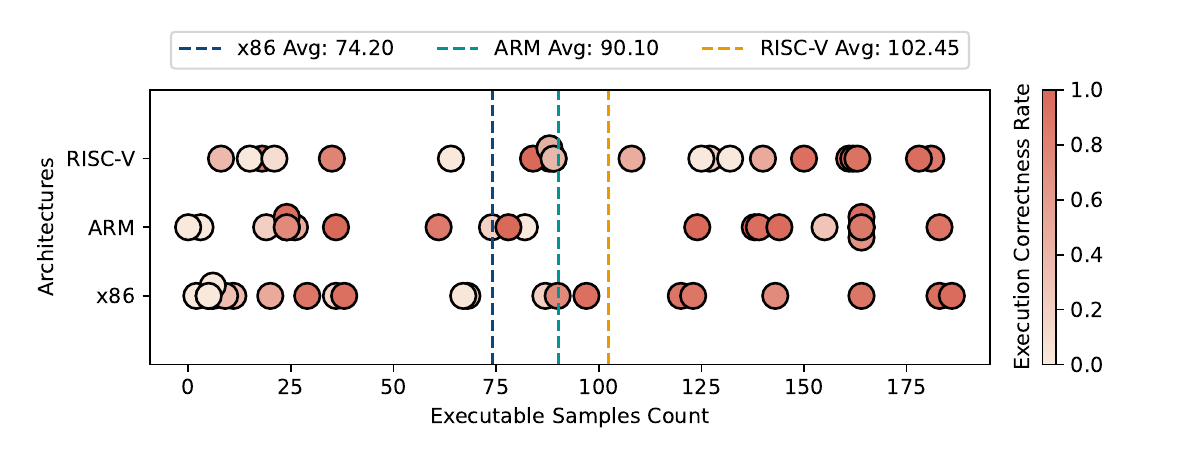}
    \caption{Evaluation of Cross-Platform Assembly Code Generation Performance for Claude-3.5-Sonnet.}
    \label{fig:platforms}
\end{figure}

Furthermore, we evaluated the quality of assembly code generation across architectures by analyzing the executable samples count and their correctness rates. As shown in Figure \ref{fig:platforms}, the position of each point represents the number of executable samples for the target architecture. The data for each row corresponds to 20 cases from the CompilerEval dataset. Points positioned further right indicate fewer syntax errors in the generated code, enabling successful executable generation. The dashed line in Figure \ref{fig:platforms} shows that RISC-V achieved the highest average number of executable samples, followed by ARM, with x86 showing the lowest count. The color intensity of scatter points reflects execution accuracy, with darker shades representing higher executable samples correctness rates. 
ARM and RISC-V architectures exhibit more dark-colored points toward the right compared to x86, indicating Claude's assembly code generation for ARM and RISC-V outperforms that for x86.

% 总的来说，Claude 这类 LLM 展现出生成跨平台汇编代码的基本能力，但在不同硬件架构上的表现存在明显差异。实验数据表明，ARM 和 RISC-V 架构的代码生成质量显著优于 x86。我们认为这主要归因于 ARM 和 RISC-V 作为精简指令集的特点，其指令相对简单，更易于 LLM 理解和生成。相比之下，生成 x86 这类复杂指令集的代码时，在访存指令和操作数使用上经常出现错误。不过，精简指令集的使用也带来了代码长度增加的问题，导致某些情况下因上下文长度限制而无法完成完整的汇编代码生成。

In summary, LLMs like Claude demonstrate the capability to generate cross-platform assembly code, but the effectiveness varies across hardware architectures. 
Experimental data shows that code generation quality for ARM and RISC-V is superior to x86. 
We attribute this result to the characteristics of ARM and RISC-V as reduced instruction set architectures, where instructions are relatively simple for LLMs to understand and generate. In contrast, generating code for complex instruction set architectures like x86 poses more significant challenges, particularly in memory access instructions and register handling. However, the use of reduced instruction sets also brings the challenge of increased code length, which sometimes prevents complete assembly code generation due to context length limitations.

\section{Design Concepts and Research Directions}

% 本章节根据上述分析和结论提出未来 LLM As A Compiler 框架的构想，并且提出该研究方向的挑战与研究点。

Based on the analysis and conclusions above, this section presents the vision for the future LaaC framework, highlighting challenges and research directions associated with this paradigm.

\subsection{Design Concepts of LaaC}

% LLM As A Compiler 的框架设计构想主要聚焦于三部分：（1）目标代码生成的基本能力（2）代码分析及优化能力（3）目标代码的纠错能力.

The framework design concepts of LaaC primarily focus on three key aspects: (1) target code generation capabilities, (2) code analysis and optimization capabilities, and (3) debugging capabilities. The detailed component structure is illustrated in Figure \ref{fig:overview}.

% 目标代码生成的基本能力由 LLM、 Prompt 和 知识库 共同负责。基于前文实验得出的结论，LLM 具备作为编译器将源代码生成汇编代码的基本能力，提供 prompt 支持可以提高其汇编代码生成的准确率。因此，在我们的框架设计中， LLM 学习编译器的映射本质，prompt 针对 LLM 进行辅助，知识库中包含源语言和硬件指令的信息。这样设计的好处是，LLM 只需要进行一次训练就可以支持各种编程语言和硬件指令。新的语言和硬件平台的引入，只需要添加 prompt 和 知识库 即可，无需承担重复训练 LLM 的开销。

\textbf{The target code generation capabilities are driven by the LLM, prompts, and the knowledge base.} In the LaaC framework, the LLM grasps the fundamental code transformation principles, the knowledge base stores information about source languages and hardware instruction sets, and prompts guide the code generation process. 
A key advantage of this architecture is that the LLM only requires a single training phase to support multiple programming languages and hardware architectures. 
Therefore, adding support for new languages or hardware platforms becomes a straightforward process of updating prompts and knowledge bases, eliminating the need for costly LLM retraining.

% 代码分析和优化能力有 reasoning engine 负责。基于前文结论，带有 reasoning 能力的 GPT-o1-mini 在代码生成效果上优于 GPT-4o 模型。因此，本文认为，LLM As A Compiler 框架可以使用 reasoning 步进推理的能力来模拟传统编译器的代码分析和变换以及。以及可以使用传统编译器的代码分析和优化策略来训练 reasoning 组件，使其具备理解程序逻辑以及调用关系的推理能力，并且可以针对性地进行代码分析，最终生成优化后的汇编代码。

\textbf{The reasoning methods can serve as the cornerstone for code analysis and optimization capabilities.} 
Inspired by the findings in Figure \ref{fig:compare}-(c) and the chain-of-thought idea \cite{wei2022chain}, we propose that the LaaC framework can leverage step-by-step reasoning to simulate the code analysis and transformations of traditional compilers. 
Additionally, traditional compiler code analysis and optimization strategies can be used to train the reasoning component, enabling it to understand program logic, perform targeted code analysis, and ultimately generate optimized assembly code.

% 目标代码的纠错能力需要针对性地设计 debug 机制。基于前文结论，LLM 具备生成正确汇编代码的能力，并且在使用多个结果验证通过率的情况下效果很可观。因此，本文认为，LLM 短期内虽难以实现生成一次汇编代码即正确的能力，但是通过目标代码的纠错能力可以实现其。例如，debug 机制可以通过生成多个结果并且将报错信息迭代反馈给 LLM 的方式最终生成正确代码，或者将报错位置和信息进行分析形成详细的报错信息指导用户进行修改。

\textbf{The debugging capabilities of LaaC can be enhanced through specialized LLM-based mechanisms.} 
By embedding relevant debug information directly into the assembly code, LaaC can integrate with existing debuggers. 
More importantly, LaaC leverages LLM-driven approaches to gain a deep understanding of source programs, combining this insight with information gathered during compilation to provide more detailed and precise debugging guidance. Through the integration of LLMs, prompts, knowledge bases, and reasoning techniques, the LLM-powered debugging mechanism can provide developers with better troubleshooting support.

\subsection{Research Directions of LaaC}

% \subsubsection{How to train a LLM that meet compiler scenario constraints?}

% 为了使 LLM as a compiler 在实际应用中可行，我们总结了三个关键约束：编译准确性、编译开销和上下文长度。编译准确性是最基础的指标。即使使用当前最先进的 LLM，其准确率仍未达到实用标准，这表明需要针对编译场景进行专门的模型精调。值得注意的是，传统的模型缩放定律（scaling law）可能并不适用于编译场景，因为编译开销同样是一个重要约束。在编译开销主要是指时间开销和编译成本。目前基于 LLM 的编译方案在执行时间上远逊于传统编译器，同时其计算成本也远超传统方案。作为基础软件设施，这样的开销水平是不可接受的。此外，当前主流 LLM（如 GPT-4）的上下文长度限制也是一个显著问题。仅能处理数百行代码的能力无法满足大型软件项目的编译需求。因此，未来的研究需要在模型架构上进行创新，以支持更长的上下文窗口。总的来说，针对编译场景的 LLM 训练出了要有足够的训练数据，更关键的是开发出能同时满足这三个核心约束的模型架构。

\textbf{(1) How can LaaC train an LLM that meets compilation scenario constraints?}
To make LaaC feasible for practical use, we have identified three key constraints: compilation accuracy, compilation cost, and context length.  

Compilation accuracy is the most fundamental metric. Even with current state-of-the-art LLMs, their accuracy still falls short of practical standards, highlighting the need for specialized fine-tuning tailored to compilation scenarios. Furthermore, leveraging reasoning methods to establish compilation-specific chain-of-thought is also essential for enhancing accuracy. 

The compilation cost of LaaC primarily involves time and computational costs. LLMs evaluated in this paper require significantly more compilation time and computational resources than traditional compilers. Therefore, optimizing fine-tuned models via knowledge distillation is key to ensuring LaaC achieves efficient response times and performance while maintaining accuracy similar to larger models.

The context length limitations of current mainstream LLMs restrict their ability to handle large programs, especially when generating code for reduced instruction set architectures. Future research should focus on developing long-context support mechanisms for compilation scenarios. For instance, sparse attention mechanisms \cite{beltagy2020longformer} could help reduce the computational complexity of long contexts, while memory modules \cite{sukhbaatar2015end} might be introduced to store context information more efficiently.

% In summary, LLM training for compilation tasks requires not only sufficient training data but, more importantly, the development of model architectures that can simultaneously address these three core constraints.

% 为了避免 LaaC 支持多种语言和硬件平台的时候重复训练，我们建议让 LLM 专注于学习与具体语言和硬件平台无关的编译核心原理，同时开发相关基础设施来对接各种编程语言和硬件架构。

\textbf{(2) How can LaaC effectively bridge diverse programming languages and multiple hardware instruction sets?}
To avoid retraining LaaC for each new language and hardware platform, we propose that the LLM focus on learning core compilation principles, while developing the related infrastructure to interface with different programming languages and hardware architectures.

% 编译过程可以构建成 Chain of Thought 形成 reasoning method。并且结合图 X-(3) 的实验结果，仿照 GPT-o1 的思路使用强化学习来对编译场景进行精细化微调。通过将 CoT 与强化学习相结合，可以将编译过程的步进式策略有效地整合到端到端的推理流程中。

The compilation process can be structured as a chain-of-thought to establish a reasoning method. Additionally, building on the experimental results in Figure \ref{fig:compare}-(c), the method can be designed to follow GPT-o1's strategy of employing reinforcement learning for fine-tuning in compilation scenarios. By integrating chain-of-thought with reinforcement learning, step-by-step compilation strategies can be effectively integrated into the end-to-end inference process.

% 编程语言语法规范和硬件平台指令集可以通过知识库存储，并结合 prompt engineering 来实现多对多的灵活映射。这种方法可以避免重复训练，但同时也带来了新的挑战：知识库和prompt信息会占用宝贵的上下文长度。因此，一个重要的研究方向是设计高效的知识压缩格式，在保证信息完整性的同时优化与LLM的交互效率。

Specifications of programming languages and hardware instruction sets can be stored in a knowledge base and combined with prompt engineering to enable flexible many-to-many mappings. While this approach eliminates the need for redundant training, it introduces new challenges, notably the consumption of valuable context length by the knowledge base and prompts. As a result, a key research direction lies in designing efficient knowledge compression formats that optimize interactions with the LLM while preserving necessary information.

% % 综上所述，我们认为编程语言和硬件指令知识的高效封装，以及如何让LLM更好地理解和执行编译过程，是两个极具价值的研究方向。

% In summary, we identify two crucial research directions: efficiently encapsulating programming language and hardware instruction knowledge, and improving LLMs' comprehension and execution of the compilation process.

% 生成调试信息是编译器的一项重要功能。我们认为 LaaC 不仅需要能够生成精确的调试指令，更应该充分利用 LLM 的优势，构建更加智能和高效的调试基础设施。

\textbf{(3) How can LaaC collaborate effectively with debuggers?}
Generating debug information is a critical function of compilers. We propose that LaaC should not only generate accurate debugging instructions but also fully leverage the strengths of LLMs to create smarter and more efficient debugging infrastructure.

% 生成准确的调试信息是 LaaC 面临的一个重要研究课题。编译器需要在编译过程中生成符号表和调试信息，这些信息不仅帮助调试器建立机器码和源代码之间的映射关系，也使开发者能够通过断点机制检查程序状态。考虑到 LLM 在生成精确代码方面的挑战性，LaaC 需要探索如何将调试指令的生成有效地整合到 LLM 的训练过程和推理方法中。

Generating accurate debugging information is a critical challenge for LaaC. During compilation, compilers produce symbol tables and debugging information that help debuggers map machine code to source code and enable developers to inspect program states through breakpoints. 
Since LLMs struggle to generate accurate assembly code, LaaC needs to explore ways to effectively integrate debugging instruction generation into LLM training and reasoning processes.

% 设计 AI 驱动的 debugger 与 LaaC 协同工作是一个极具创新性的研究方向。LLM 强大的代码理解与生成能力已在 GitHub Copilot 等工具中得到验证。将这种能力扩展到调试场景，并与 LaaC 深度整合，我们可以构建新一代智能调试基础设施。这种 AI 驱动的调试解决方案可以生成更加详细的调试报告，甚至能够自动修复代码缺陷，相比传统的断点调试方法将大大提升开发效率。

Designing an LLM-driven debugger to complement LaaC is an innovative research direction. LLMs have already demonstrated their strong code comprehension and generation abilities through tools like GitHub Copilot. By extending these capabilities to debugging and integrating them with LaaC, we can create an intelligent debugging infrastructure. This LLM-driven solution could generate more detailed debugging reports and even automatically fix code defects, significantly enhancing development efficiency compared to traditional breakpoint-based methods.

% % 总之，无论是使用传统 debugger 还是设计 AI 驱动的 debugger，关键问题主要是编译器和调试器的协同方式需要准确并高效地定位程序问题并解决问题。

% In summary, whether using a traditional debugger or designing an AI-driven debugger, the key challenge lies in ensuring effective collaboration between LaaC and the debugger to precisely and swiftly identify and fix program issues.

\section{Related Work}

% 本章节将从两个主要角度探讨相关工作：首先分析 LLM 在编译器领域的研究进展，其次讨论 LLM 数据集评估指标的应用。我们将探讨这些现有研究与本文工作的区别与联系。

This section will discuss related work from two main perspectives: the research progress of LLMs for the compiler domain and the evaluation metrics used in the LLM evaluation dataset.

\subsection{Related Work of LLMs in Compiler Technology}

% 近年来，LLM在编译器领域的应用是一个热门的研究话题。在开发方面，研究人员利用LLM和专门数据集来辅助编译器后端开发；在优化领域，开发了多种基于LLM的方法来提升代码生成质量，包括执行速度和代码体积的优化；在验证方面，LLM被用于辅助确保编译程序的正确性和可靠性；在测试领域，LLM可以生成测试程序来检测潜在的深层逻辑错误。此外，在反汇编领域，LLM展现出了解决汇编代码信息密度低、变量名恢复困难、可读性差等问题的能力。

Applying LLMs in compiler development and maintenance has become a significant research direction in recent years. In the development domain, researchers have leveraged LLMs and specialized datasets to assist in developing compiler backends \cite{zhongcomback}. In the optimization domain, LLM-based methods \cite{cummins2023large,cummins2024meta,hong2024llm} have been developed to improve code generation quality. In the verification domain, LLMs have been used to help ensure the correctness and reliability of compiled programs \cite{munley2024llm4vv}. In the testing domain, LLMs can generate test programs to detect potential deep logical errors \cite{yang2024whitefox}. Furthermore, in the decompilation domain, LLMs \cite{tan2024llm4decompile,xu2023lmpa,armengol2024slade} have demonstrated the ability to address issues such as low information density in assembly code, difficulties in recovering variable names, and poor readability.

% 与这些将LLM作为辅助工具的研究不同，本文致力于验证 LLM 直接作为端到端编译器的可能性。这种方法可以显著降低编译器各模块的开发与维护开销。尽管早期使用transformer模型进行C代码到x86汇编转换的研究并未得出乐观结论，但本文通过更全面的实验评估和更多样的模型与硬件平台测试，展示了LLM作为编译器的巨大潜力。同时，我们也提出了未来的框架设计和明确的研究方向。

In contrast to the above researches that use LLMs as supporting tools within the compilation pipeline, this paper explores the feasibility of utilizing end-to-end LLM technology directly as a compiler. 
This approach has the potential to reduce the development and maintenance overhead of individual compiler modules. 
Although the early study \cite{armengol2021learning} using transformer models for C to x86 assembly translation did not yield optimistic results, our paper demonstrates the potential of LLMs to generate assembly code through more comprehensive experimental evaluations and diverse tests across multiple models and hardware platforms. Additionally, we propose a future framework design and outline clear directions for further research.

\subsection{Evaluation Metrics in LLM Datasets}

% 在 LLM 数据集的相关工作中，评估指标的选择对于衡量模型性能至关重要。常见的评估指标包括：任务通过率(pass@k)，用于衡量在给定样本中选取k个方案时至少有一个方案能够通过的概率；精确匹配准确率(EM)，用于测量模型生成结果与真实结果完全一致的比例；编辑距离相似度(Levenshtein Edit Distance)，通过计算将一个字符串转换为另一个所需的最少编辑操作来量化相似程度；以及BLEU评分，这是一种基于n-gram的评估方法，主要用于衡量生成文本与参考文本在结构和语法模式上的相似度。

In related work of LLM datasets \cite{kulal2019spoc,chen2021evaluating,lu2021codexglue,zhongcomback}, various metrics have been employed to evaluate the performance of LLMs.
Pass rate (\texttt{pass@k}) measures the probability that at least one solution among \texttt{k} selected samples passes a predefined correctness check.
Exact Match (EM) accuracy measures the proportion of model-generated results that perfectly match the ground truth.
Levenshtein Edit Distance (ED) similarity quantifies similarity by calculating the minimum number of editing operations needed to transform one string into another.
BLEU score is an n-gram-based evaluation method primarily used to measure the structural and grammatical similarity between generated text and reference text.

% 在汇编代码级别，相同的功能可以通过不同的指令组合实现，这也是指令调度的意义所在。在实际应用中，修改 LLM 生成的汇编代码也是无意义的。因此，我们认为上述文本级别的准确率和相似度评估指标 EM, ED, BLEU 并不适合作为本文的评估指标。对于 LaaC 来说，最重要的指标是编译结果词法、语法、语义的正确率。本文参考 HumanEval pass@k 和 Spoc 的 success rate 针对编译场景提出了 Compilation Success Rate，Executable Samples Count，Executable Samples Correctness Rate 来描述 LLMs 生成汇编代码的正确性。

% 在评估 LLM 生成的汇编代码方面，我们认为上述文本级别的准确率和相似度 EM, ED, BLEU 并不适合作为评估指标。
% 在汇编代码级别，相同的功能可以通过不同的指令组合实现，这也是指令调度的意义所在。
% 因此，比较汇编代码的相似度是无意义的。
% 对于 LaaC 来说，最重要的指标是编译结果词法、语法、语义的正确率。

When evaluating assembly code generated by LLMs, the text-level accuracy and similarity metrics, such as EM, ED, and BLEU, are not suitable. At the assembly level, identical functionality can be achieved through different combinations of instructions due to variations in instruction scheduling and optimization strategies. Therefore, comparing code similarity at the assembly level is meaningless. 
% At the assembly code level, the same functionality can be achieved through different combinations of instructions, highlighting the importance of instruction scheduling. In practical applications, modifying assembly code generated by LLMs is not meaningful. Therefore, the above text-level accuracy and similarity evaluation metrics, such as EM, ED, and BLEU, are unsuitable for this paper. 
For LaaC, the appropriate metrics are the lexical, syntactic, and semantic accuracy of the compilation results. Drawing inspiration from the \texttt{pass@k} of HumanEval \cite{chen2021evaluating} and the success rate of SpoC \cite{kulal2019spoc}, this paper proposes three metrics for compilation scenarios: \textit{Compilation Success Rate}, \textit{Executable Samples Count}, and \textit{Executable Samples Correctness Rate}, which aim to describe the correctness of assembly code generated by LLMs.

\section{Conclusion}

% 本研究开创性地探索了将大语言模型(LLM)作为端到端编译器的可能性，通过构建 CompilerEval 评估框架和测试集，系统地验证了 LLM 在编译领域的应用潜力。我们从代码生成通过率、模型类型与规模、高性能代码生成等多个维度进行了全面评估，不仅证实了 LLM 作为编译器的可行性，同时提出了 LLM As A Compiler 框架的关键组件，并阐述了这一研究方向的关键挑战与研究重点。我们相信 LLM 驱动的编译方法能够带来编程语言与编译技术的革命性范式转移。

This paper explores the feasibility of using end-to-end LLM technology as a compiler. We assess the ability of LLMs to generate assembly code by creating the CompilerEval dataset and framework. Our evaluation focuses on the limitations of the LLMs in generating assembly code, the strategies to enhance the compilation success rate, and the ability to generate cross-platform assembly code. Experimental results show that current mainstream LLMs with the appropriate mechanisms can produce the correct assembly code for simple kernels. However, the overall compilation success rate is still low. This indicates that while LLM-based methods demonstrate potential as compilers, further research is needed to improve the success rate to meet practical standards. Based on our findings, we propose the core components of the LaaC framework and discuss the key challenges and research areas in this field. We believe that end-to-end LLM approaches could lead to a paradigm shift in compiler technology.

\bibliographystyle{IEEEtran}
\bibliography{ref}

@INPROCEEDINGS{MLIR,
  author={Lattner, Chris and Amini, Mehdi and Bondhugula, Uday and Cohen, Albert and Davis, Andy and Pienaar, Jacques and Riddle, River and Shpeisman, Tatiana and Vasilache, Nicolas and Zinenko, Oleksandr},
  booktitle={2021 IEEE/ACM International Symposium on Code Generation and Optimization (CGO)}, 
  title={MLIR: Scaling Compiler Infrastructure for Domain Specific Computation}, 
  year={2021},
  volume={},
  number={},
  pages={2-14},
  doi={10.1109/CGO51591.2021.9370308}
}

@misc{GPT-4o,
  author = {OpenAI},
  title = {Hello gpt-4o},
  url = {https://openai.com/index/hello-gpt-4o/},
  note = {Accessed: 2025-01}
}

@misc{GPT-o1,
  author = {OpenAI},
  title = {Introducing OpenAI o1},
  url = {https://openai.com/o1/},
  note = {Accessed: 2025-01}
}

@misc{Claude-3.5,
  author = {Anthropic},
  title = {Claude 3.5 Sonnet},
  url = {https://www.anthropic.com/claude/sonnet},
  note = {Accessed: 2025-01}
}

@misc{Gemini-2.0,
  author = {Google DeepMind},
  title = {Gemini 2.0 Flash Experimental},
  url = {https://deepmind.google/technologies/gemini/flash/},
  note = {Accessed: 2025-01}
}

@article{dubey2024llama,
  title={The llama 3 herd of models},
  author={Dubey, Abhimanyu and Jauhri, Abhinav and Pandey, Abhinav and Kadian, Abhishek and Al-Dahle, Ahmad and Letman, Aiesha and Mathur, Akhil and Schelten, Alan and Yang, Amy and Fan, Angela and others},
  journal={arXiv preprint arXiv:2407.21783},
  year={2024}
}

@book{alfred2007compilers,
  title={Compilers principles, techniques \& tools},
  author={Alfred, V Aho and Monica, S Lam and Jeffrey, D Ullman},
  year={2007},
  publisher={pearson Education}
}

@book{andrew2004modern,
  title={Modern Compiler Implementation in C},
  author={Andrew W. Appel and Maia Ginsburg},
  year={2004},
  publisher={Cambridge University Press}
}

@book{muchnick1997advanced,
  title={Advanced compiler design implementation},
  author={Muchnick, Steven},
  year={1997},
  publisher={Morgan kaufmann}
}

@inproceedings{lattner2004llvm,
  title={LLVM: A compilation framework for lifelong program analysis \& transformation},
  author={Lattner, Chris and Adve, Vikram},
  booktitle={International symposium on code generation and optimization, 2004. CGO 2004.},
  pages={75--86},
  year={2004},
  organization={IEEE}
}

@misc{gcc,
  author = {GNU Project},
  title = {GCC, the GNU Compiler Collection},
  url = {https://gcc.gnu.org/}
}

@inproceedings{yang2023llm4drive,
  title={Llm4drive: A survey of large language models for autonomous driving},
  author={Yang, Zhenjie and Jia, Xiaosong and Li, Hongyang and Yan, Junchi},
  booktitle={NeurIPS 2024 Workshop on Open-World Agents},
  year={2023}
}

@inproceedings{ling2024adapting,
  title={Adapting large language model with speech for fully formatted end-to-end speech recognition},
  author={Ling, Shaoshi and Hu, Yuxuan and Qian, Shuangbei and Ye, Guoli and Qian, Yao and Gong, Yifan and Lin, Ed and Zeng, Michael},
  booktitle={ICASSP 2024-2024 IEEE International Conference on Acoustics, Speech and Signal Processing (ICASSP)},
  pages={11046--11050},
  year={2024},
  organization={IEEE}
}

@inproceedings{zhongcomback,
  title={ComBack: A Versatile Dataset for Enhancing Compiler Backend Development Efficiency},
  author={Zhong, Ming and Lyu, Fang and Wang, Lulin and Geng, Hongna and Qiu, Lei and Cui, Huimin and Feng, Xiaobing},
  booktitle={The Thirty-eight Conference on Neural Information Processing Systems Datasets and Benchmarks Track}
}

@inproceedings{lakomkin2024end,
  title={End-to-end speech recognition contextualization with large language models},
  author={Lakomkin, Egor and Wu, Chunyang and Fathullah, Yassir and Kalinli, Ozlem and Seltzer, Michael L and Fuegen, Christian},
  booktitle={ICASSP 2024-2024 IEEE International Conference on Acoustics, Speech and Signal Processing (ICASSP)},
  pages={12406--12410},
  year={2024},
  organization={IEEE}
}

@article{chen2021evaluating,
  title={Evaluating large language models trained on code},
  author={Chen, Mark and Tworek, Jerry and Jun, Heewoo and Yuan, Qiming and Pinto, Henrique Ponde De Oliveira and Kaplan, Jared and Edwards, Harri and Burda, Yuri and Joseph, Nicholas and Brockman, Greg and others},
  journal={arXiv preprint arXiv:2107.03374},
  year={2021}
}

@inproceedings{jin2023inferfix,
  title={Inferfix: End-to-end program repair with llms},
  author={Jin, Matthew and Shahriar, Syed and Tufano, Michele and Shi, Xin and Lu, Shuai and Sundaresan, Neel and Svyatkovskiy, Alexey},
  booktitle={Proceedings of the 31st ACM Joint European Software Engineering Conference and Symposium on the Foundations of Software Engineering},
  pages={1646--1656},
  year={2023}
}

@article{cummins2023large,
  title={Large language models for compiler optimization},
  author={Cummins, Chris and Seeker, Volker and Grubisic, Dejan and Elhoushi, Mostafa and Liang, Youwei and Roziere, Baptiste and Gehring, Jonas and Gloeckle, Fabian and Hazelwood, Kim and Synnaeve, Gabriel and others},
  journal={arXiv preprint arXiv:2309.07062},
  year={2023}
}

@article{armengol2021learning,
  title={Learning c to x86 translation: An experiment in neural compilation},
  author={Armengol-Estap{\'e}, Jordi and O'Boyle, Michael FP},
  journal={arXiv preprint arXiv:2108.07639},
  year={2021}
}

@article{tan2024llm4decompile,
  title={LLM4Decompile: Decompiling Binary Code with Large Language Models},
  author={Tan, Hanzhuo and Luo, Qi and Li, Jing and Zhang, Yuqun},
  journal={arXiv preprint arXiv:2403.05286},
  year={2024}
}

@article{xu2023lmpa,
  title={Lmpa: Improving decompilation by synergy of large language model and program analysis},
  author={Xu, Xiangzhe and Zhang, Zhuo and Feng, Shiwei and Ye, Yapeng and Su, Zian and Jiang, Nan and Cheng, Siyuan and Tan, Lin and Zhang, Xiangyu},
  journal={arXiv preprint arXiv:2306.02546},
  year={2023}
}

@inproceedings{armengol2024slade,
  title={SLaDe: A Portable Small Language Model Decompiler for Optimized Assembly},
  author={Armengol-Estap{\'e}, Jordi and Woodruff, Jackson and Cummins, Chris and O'Boyle, Michael FP},
  booktitle={2024 IEEE/ACM International Symposium on Code Generation and Optimization (CGO)},
  pages={67--80},
  year={2024},
  organization={IEEE}
}

@article{kulal2019spoc,
  title={Spoc: Search-based pseudocode to code},
  author={Kulal, Sumith and Pasupat, Panupong and Chandra, Kartik and Lee, Mina and Padon, Oded and Aiken, Alex and Liang, Percy S},
  journal={Advances in Neural Information Processing Systems},
  volume={32},
  year={2019}
}

@article{wang2024visionllm,
  title={Visionllm: Large language model is also an open-ended decoder for vision-centric tasks},
  author={Wang, Wenhai and Chen, Zhe and Chen, Xiaokang and Wu, Jiannan and Zhu, Xizhou and Zeng, Gang and Luo, Ping and Lu, Tong and Zhou, Jie and Qiao, Yu and others},
  journal={Advances in Neural Information Processing Systems},
  volume={36},
  year={2024}
}

@article{hennessy2019new,
  title={A new golden age for computer architecture},
  author={Hennessy, John L and Patterson, David A},
  journal={Communications of the ACM},
  volume={62},
  number={2},
  pages={48--60},
  year={2019},
  publisher={ACM New York, NY, USA}
}

@inproceedings{lattner2023golden,
  title={The golden age of compiler design in an era of HW/SW co-design},
  author={Lattner, Chris},
  year={2023},
  organization={ASPLOS Technical Report, 2021.}
}

@article{zhang2023compiler,
  title={Compiler Technologies in Deep Learning Co-Design: A Survey},
  author={Zhang, Hongbin and Xing, Mingjie and Wu, Yanjun and Zhao, Chen},
  journal={Intelligent Computing},
  volume={2},
  pages={0040},
  year={2023},
  publisher={AAAS}
}

@inproceedings{chen2018tvm,
  title={$\{$TVM$\}$: An automated $\{$End-to-End$\}$ optimizing compiler for deep learning},
  author={Chen, Tianqi and Moreau, Thierry and Jiang, Ziheng and Zheng, Lianmin and Yan, Eddie and Shen, Haichen and Cowan, Meghan and Wang, Leyuan and Hu, Yuwei and Ceze, Luis and others},
  booktitle={13th USENIX Symposium on Operating Systems Design and Implementation (OSDI 18)},
  pages={578--594},
  year={2018}
}

@article{grossman2023compile,
  title={Compile: A large ir dataset from production sources},
  author={Grossman, Aiden and Paehler, Ludger and Parasyris, Konstantinos and Ben-Nun, Tal and Hegna, Jacob and Moses, William and Diaz, Jose M Monsalve and Trofin, Mircea and Doerfert, Johannes},
  journal={arXiv preprint arXiv:2309.15432},
  year={2023}
}

@inproceedings{zhai2023tlp,
  title={Tlp: A deep learning-based cost model for tensor program tuning},
  author={Zhai, Yi and Zhang, Yu and Liu, Shuo and Chu, Xiaomeng and Peng, Jie and Ji, Jianmin and Zhang, Yanyong},
  booktitle={Proceedings of the 28th ACM International Conference on Architectural Support for Programming Languages and Operating Systems, Volume 2},
  pages={833--845},
  year={2023}
}

@inproceedings{mendis2019ithemal,
  title={Ithemal: Accurate, portable and fast basic block throughput estimation using deep neural networks},
  author={Mendis, Charith and Renda, Alex and Amarasinghe, Saman and Carbin, Michael},
  booktitle={International Conference on machine learning},
  pages={4505--4515},
  year={2019},
  organization={PMLR}
}

@inproceedings{turner2021neural,
  title={Neural architecture search as program transformation exploration},
  author={Turner, Jack and Crowley, Elliot J and O'Boyle, Michael FP},
  booktitle={Proceedings of the 26th ACM International Conference on Architectural Support for Programming Languages and Operating Systems},
  pages={915--927},
  year={2021}
}

@inproceedings{park2022srtuner,
  title={SRTuner: Effective compiler optimization customization by exposing synergistic relations},
  author={Park, Sunghyun and Latifi, Salar and Park, Yongjun and Behroozi, Armand and Jeon, Byungsoo and Mahlke, Scott},
  booktitle={2022 IEEE/ACM International Symposium on Code Generation and Optimization (CGO)},
  pages={118--130},
  year={2022},
  organization={IEEE}
}

@article{chakraborty2023ranking,
  title={Ranking llm-generated loop invariants for program verification},
  author={Chakraborty, Saikat and Lahiri, Shuvendu K and Fakhoury, Sarah and Musuvathi, Madanlal and Lal, Akash and Rastogi, Aseem and Senthilnathan, Aditya and Sharma, Rahul and Swamy, Nikhil},
  journal={arXiv preprint arXiv:2310.09342},
  year={2023}
}

@article{munley2024llm4vv,
  title={LLM4VV: Developing LLM-driven testsuite for compiler validation},
  author={Munley, Christian and Jarmusch, Aaron and Chandrasekaran, Sunita},
  journal={Future Generation Computer Systems},
  year={2024},
  publisher={Elsevier}
}

@article{yang2024whitefox,
  title={Whitefox: White-box compiler fuzzing empowered by large language models},
  author={Yang, Chenyuan and Deng, Yinlin and Lu, Runyu and Yao, Jiayi and Liu, Jiawei and Jabbarvand, Reyhaneh and Zhang, Lingming},
  journal={Proceedings of the ACM on Programming Languages},
  volume={8},
  number={OOPSLA2},
  pages={709--735},
  year={2024},
  publisher={ACM New York, NY, USA}
}

@article{wei2022chain,
  title={Chain-of-thought prompting elicits reasoning in large language models},
  author={Wei, Jason and Wang, Xuezhi and Schuurmans, Dale and Bosma, Maarten and Xia, Fei and Chi, Ed and Le, Quoc V and Zhou, Denny and others},
  journal={Advances in neural information processing systems},
  volume={35},
  pages={24824--24837},
  year={2022}
}

@article{beltagy2020longformer,
  title={Longformer: The long-document transformer},
  author={Beltagy, Iz and Peters, Matthew E and Cohan, Arman},
  journal={arXiv preprint arXiv:2004.05150},
  year={2020}
}

@article{sukhbaatar2015end,
  title={End-to-end memory networks},
  author={Sukhbaatar, Sainbayar and Weston, Jason and Fergus, Rob and others},
  journal={Advances in neural information processing systems},
  volume={28},
  year={2015}
}

@article{parr1995antlr,
  title={ANTLR: A predicated-LL (k) parser generator},
  author={Parr, Terence J. and Quong, Russell W.},
  journal={Software: Practice and Experience},
  volume={25},
  number={7},
  pages={789--810},
  year={1995},
  publisher={Wiley Online Library}
}

@inproceedings{hong2024llm,
  title={LLM-Aided Compilation for Tensor Accelerators},
  author={Hong, Charles and Bhatia, Sahil and Haan, Altan and Dong, Shengjun Kris and Nikiforov, Dima and Cheung, Alvin and Shao, Yakun Sophia},
  booktitle={2024 IEEE LLM Aided Design Workshop (LAD)},
  pages={1--14},
  year={2024},
  organization={IEEE}
}

@article{cummins2024meta,
  title={Meta large language model compiler: Foundation models of compiler optimization},
  author={Cummins, Chris and Seeker, Volker and Grubisic, Dejan and Roziere, Baptiste and Gehring, Jonas and Synnaeve, Gabriel and Leather, Hugh},
  journal={arXiv preprint arXiv:2407.02524},
  year={2024}
}

@article{lu2021codexglue,
  title={Codexglue: A machine learning benchmark dataset for code understanding and generation},
  author={Lu, Shuai and Guo, Daya and Ren, Shuo and Huang, Junjie and Svyatkovskiy, Alexey and Blanco, Ambrosio and Clement, Colin and Drain, Dawn and Jiang, Daxin and Tang, Duyu and others},
  journal={arXiv preprint arXiv:2102.04664},
  year={2021}
}

% \begin{thebibliography}{00}
% \bibitem{b1} G. Eason, B. Noble, and I. N. Sneddon, ``On certain integrals of Lipschitz-Hankel type involving products of Bessel functions,'' Phil. Trans. Roy. Soc. London, vol. A247, pp. 529--551, April 1955.
% \bibitem{b2} J. Clerk Maxwell, A Treatise on Electricity and Magnetism, 3rd ed., vol. 2. Oxford: Clarendon, 1892, pp.68--73.
% \bibitem{b3} I. S. Jacobs and C. P. Bean, ``Fine particles, thin films and exchange anisotropy,'' in Magnetism, vol. III, G. T. Rado and H. Suhl, Eds. New York: Academic, 1963, pp. 271--350.
% \bibitem{b4} K. Elissa, ``Title of paper if known,'' unpublished.
% \bibitem{b5} R. Nicole, ``Title of paper with only first word capitalized,'' J. Name Stand. Abbrev., in press.
% \bibitem{b6} Y. Yorozu, M. Hirano, K. Oka, and Y. Tagawa, ``Electron spectroscopy studies on magneto-optical media and plastic substrate interface,'' IEEE Transl. J. Magn. Japan, vol. 2, pp. 740--741, August 1987 [Digests 9th Annual Conf. Magnetics Japan, p. 301, 1982].
% \bibitem{b7} M. Young, The Technical Writer's Handbook. Mill Valley, CA: University Science, 1989.
% \end{thebibliography}
% \vspace{12pt}

\end{document}